\documentclass[letterpaper, 10 pt, conference]{ieeeconf}  

\IEEEoverridecommandlockouts                              

\usepackage{amsmath} 
\usepackage{amssymb}  
\usepackage{nicefrac}
\usepackage{graphicx}
\usepackage{xcolor}
\usepackage{makecell}
\usepackage[font=small]{caption}
\usepackage{pifont}
\newcommand{\cmark}{\ding{51}}%
\usepackage{xifthen}
\usepackage{xparse}
\usepackage{subdepth}
\usepackage{leftidx}
\usepackage{bm}
\usepackage{etoolbox}
\usepackage{amsmath}
\usepackage{multirow, makecell}

\newcommand{\bbm}{\begin{bmatrix}}
\newcommand{\ebm}{\end{bmatrix}}

\DeclareMathAlphabet{\mybf}{OT1}{ptm}{b}{n} 
\newcommand{\mybs}[1]{{\bm{#1}}} 
\DeclareMathAlphabet{\mybfi}{OML}{cmm}{b}{it}

\newcommand{\mbf}[1]{
\ifcat\noexpand#1\relax 
\mybs{#1}
\else
\mybf{#1}
\fi
}

\newcommand{\mbfbar}[1]{{\overline{\mbf{#1}}}}
\newcommand{\mbfhat}[1]{{\hat{\mbf{#1}}}}
\newcommand{\mbftilde}[1]{{\tilde{\mbf{#1}}}}

\newcommand{\mbfdot}[1]{{\dot {\mbf{#1}}}}

\NewDocumentCommand{\mbfidentity}{o}{\IfValueTF{#1}{\mbf{I}_{#1\hspace{\rightshift}}}{\mbf{I}}}
\NewDocumentCommand{\mbfzero}{oo}{\IfValueTF{#1}{\mbf{0}_{#1\times#2\hspace{\rightshift}}}{\mbf{0}}}

\newcommand{\cframe}[1]{{\smash{\protect\underrightarrow{\mathcal{F}}_{#1}}}}

\newcommand{\homo}[1]{{\mybfi{#1}}}

\newcommand{\mbfh}[1]{{\homo{#1}}}

\newcommand{\crossmx}[1]{\left[{#1}\right]^\times}

\newlength{\leftshift}
\newlength{\rightshift}
\newcommand{\mvec}[2]{\leftidx{_{#1}}{\mbf #2}{}} 

\newcommand{\pos}[2]{\leftidx{_{#1}}{ \mbf r}{_{#2\hspace{\rightshift}}}} 
\newcommand{\postilde}[2]{\leftidx{_{#1}}{\mbftilde r}{_{#2\hspace{\rightshift}}}} 

\NewDocumentCommand{\vel}{moo}{
	\IfValueTF{#1}{\leftidx{_{#1}}}{}{\mbf v}{\IfValueTF{#2}{_{#2#3\hspace{\rightshift}}}{}}}
\NewDocumentCommand{\veltilde}{moo}{
	\IfValueTF{#1}{\leftidx{_{#1}}}{}{\mbftilde v}{\IfValueTF{#2}{_{#2#3\hspace{\rightshift}}}{}}}
\NewDocumentCommand{\velbar}{moo}{
	\IfValueTF{#1}{\leftidx{_{#1}}}{}{\mbfbar v}{\IfValueTF{#2}{_{#2#3\hspace{\rightshift}}}{}}}
\NewDocumentCommand{\velhat}{moo}{
	\IfValueTF{#1}{\leftidx{_{#1}}}{}{\mbfhat v}{\IfValueTF{#2}{_{#2#3\hspace{\rightshift}}}{}}}
\NewDocumentCommand{\veldot}{moo}{
	\IfValueTF{#1}{\leftidx{_{#1}}}{}{\mbfdot v}{\IfValueTF{#2}{_{#2#3\hspace{\rightshift}}}{}}}

\NewDocumentCommand{\acc}{moo}{
	\IfValueTF{#1}{\leftidx{_{#1}}}{}{\mbf a}{\IfValueTF{#2}{_{#2#3\hspace{\rightshift}}}{}}}
\NewDocumentCommand{\acctilde}{moo}{
	\IfValueTF{#1}{\leftidx{_{#1}}}{}{\mbftilde a}{\IfValueTF{#2}{_{#2#3\hspace{\rightshift}}}{}}}
\NewDocumentCommand{\accbar}{moo}{
	\IfValueTF{#1}{\leftidx{_{#1}}}{}{\mbfbar a}{\IfValueTF{#2}{_{#2#3\hspace{\rightshift}}}{}}}
\NewDocumentCommand{\acchat}{moo}{
	\IfValueTF{#1}{\leftidx{_{#1}}}{}{\mbfhat a}{\IfValueTF{#2}{_{#2#3\hspace{\rightshift}}}{}}}
\NewDocumentCommand{\accdot}{moo}{
	\IfValueTF{#1}{\leftidx{_{#1}}}{}{\mbfdot a}{\IfValueTF{#2}{_{#2#3\hspace{\rightshift}}}{}}}

\NewDocumentCommand{\rotvel}{moo}{
	\IfValueTF{#1}{\leftidx{_{#1}}}{}{\mbf $\omega$}{\IfValueTF{#2}{_{#2#3\hspace{\rightshift}}}{}}}
\NewDocumentCommand{\rotveltilde}{moo}{
	\IfValueTF{#1}{\leftidx{_{#1}}}{}{\mbftilde $\omega$}{\IfValueTF{#2}{_{#2#3\hspace{\rightshift}}}{}}}
\NewDocumentCommand{\rotvelbar}{moo}{
	\IfValueTF{#1}{\leftidx{_{#1}}}{}{\mbfbar $\omega$}{\IfValueTF{#2}{_{#2#3\hspace{\rightshift}}}{}}}
\NewDocumentCommand{\rotvelhat}{moo}{
	\IfValueTF{#1}{\leftidx{_{#1}}}{}{\mbfhat $\omega$}{\IfValueTF{#2}{_{#2#3\hspace{\rightshift}}}{}}}
\NewDocumentCommand{\rotveldot}{moo}{
	\IfValueTF{#1}{\leftidx{_{#1}}}{}{\mbfdot $\omega$}{\IfValueTF{#2}{_{#2#3\hspace{\rightshift}}}{}}}

\newcommand{\C}[2]{ {\mbf C}   {_{#1#2\hspace{\rightshift}} }     } 
\newcommand{\T}[2]{{\mbfh T}{_{#1#2\hspace{\rightshift}}}} 
\newcommand{\q}[2]{{\mbf q}{_{#1#2\hspace{\rightshift}}}} 

\newcommand{\Exp}[1]{\text{Exp}\left( #1 \right)}

\title{\LARGE \bf
Tightly-Coupled LiDAR-Visual-Inertial SLAM and Large-Scale Volumetric Occupancy Mapping 
}

\author{Simon Boche$^{1}$, Sebasti\'{a}n Barbas Laina$^{1}$, Stefan Leutenegger$^{1,2,3}$
\thanks{This work was supported by the Technical University of Munich, the TUM Innovation Network CoConstruct and Leica Geosystems AG.}
\thanks{$^{1}$Smart Robotics Lab, School of Computation, Information and Technology (CIT),
Technical University of Munich, Germany.
        {\tt\small firstname.surname@tum.de}}%
\thanks{$^{2}$ Munich Institute of Robotics and Machine Intelligence (MIRMI), Technical University of Munich, Germany}%
\thanks{$^{3}$ Department of Computing, Imperial College London, UK}%
}

\begin{document}

\maketitle
\thispagestyle{empty}
\pagestyle{empty}

\begin{abstract}
Autonomous navigation is one of the key requirements for every potential application of mobile robots in the real-world. Besides high-accuracy state estimation, a suitable and globally consistent representation of the 3D environment is indispensable. 
We present a fully tightly-coupled LiDAR-Visual-Inertial SLAM system and 3D mapping framework applying local submapping strategies to achieve scalability to large-scale environments. A novel and correspondence-free, inherently probabilistic, formulation of LiDAR residuals is introduced, expressed only in terms of the occupancy fields and its respective gradients. These residuals can be added to a factor graph optimisation problem, either as frame-to-map factors for the live estimates or as map-to-map factors aligning the submaps with respect to one another. 
Experimental validation demonstrates that the approach achieves state-of-the-art pose accuracy and furthermore produces globally consistent volumetric occupancy submaps which can be directly used in downstream tasks such as navigation or exploration. 
\end{abstract}


\section{Introduction}
Robust and accurate state estimation is one of the fundamental components for autonomous navigation of robotic systems. But positioning is only part of the problem. An accurate and especially globally consistent representation of the 3D environment that the robot is operating in, is also indispensable.
Simultaneous Localisation and Mapping (SLAM) approaches fusing multiple sensor sources, such as stereo vision, Inertial Measurement Units (IMU) or Light Detection and Ranging (LiDAR) sensors, have proven to achieve accurate performance in localisation. Robustness is gained by fusing complementary sensors to overcome degrading scenarios for each of the individual sensors.
Most state-of-the-art LiDAR-Visual-Inertial (LVI) SLAM systems are representing the 3D world in terms of features or point clouds, e.g.~\cite{Vilens,r3live,lvisam,ct-icp}. Lately, also surfels have been adopted to LiDAR-based SLAM systems~\cite{Wildcat}. While these representations may prove suitable for state estimation and surface reconstruction, they cannot be directly used for downstream tasks such as robotic path planning, where an explicit representation of observed free space is desirable.

That is why, lately, there has been research in coupling SLAM and volumetric mapping. To account for potential pose drift, which would lead to degrading map accuracy and consistency, submapping approaches have recently gained interest. The idea behind that is to divide the environment into several local submaps, which will stay accurate and consistent due to locally limited drift. Different strategies on how to create these submaps and to keep global consistency across submaps have been investigated, e.g.\ in~\cite{voxgraph} or~\cite{oxfordSubmappingExtended}.

Most existing methods decouple the problems of SLAM and volumetric mapping and rearrange the relative position and orientation of all submaps with respect to each other in a loosely-coupled optimisation problem. In contrast, we aim to formulate a whole unified tightly-coupled problem to keep our 3D map representations consistent at all times which makes them suitable for usage in robotic applications.
\begin{figure}[!t]
    \centering
    \includegraphics[width=\linewidth, trim={0, 2cm, 0, 1.5cm}, clip ]{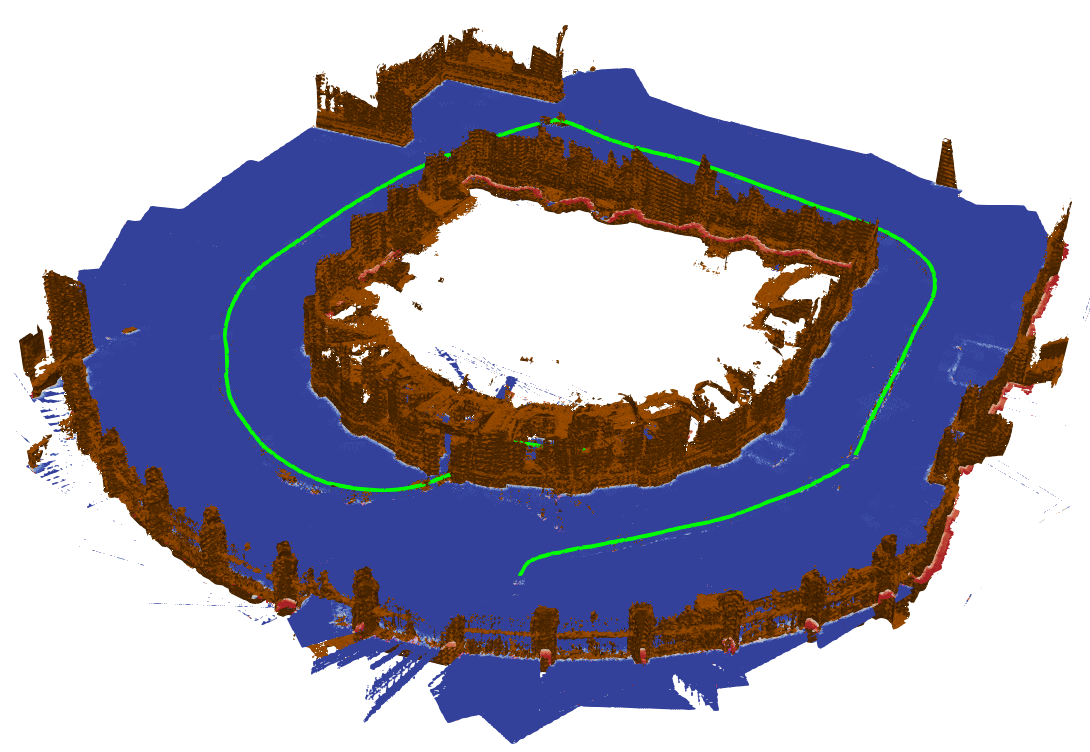}
    \caption{Horizontal slices of the occupancy fields and 3D reconstructions from Sequence \textit{Exp21} of the HILTI SLAM Challenge~\cite{Hilti22}. All submaps are overlayed. Meshes of the surfaces (brown) are extracted from the occupancy field as zero-crossings. The blue area denotes free space extracted for gravity-aligned slices through the submaps. The estimated trajectory is shown in green. }
    \vspace{-0.7cm}
    \label{fig:reconstruction}
\end{figure}
In short, the key contributions of this work are the following.
\begin{itemize}
    \item We present a tightly-coupled optimisation-based LVI-SLAM and occupancy mapping framework leveraging local submaps to ensure global consistency. 
    \item We introduce a novel residual formulation for LiDAR-based error terms that directly operates on the occupancy values and field gradients. Both can be efficiently queried from the submaps without an expensive data association step as for example in Iterative Closest Point (ICP) approaches. These residuals can be added to the factor graph in two ways, either as factors for every live frame or as factors between submaps. 
    \item We evaluate the approach quantitatively in terms of localisation accuracy and qualitatively regarding map quality. We demonstrate that our system yields globally consistent maps for different LiDAR sensors while achieving state-of-the-art localisation performance. 
\end{itemize}
The remainder of this paper is structured as follows: after a brief summary of related works in Sec.~\ref{chapter:related_work}, Sec.~\ref{chapter:preliminaries} presents the problem statement and notations. Sec.~\ref{chapter:system_overview} sketches the overall structure of the system before the mapping and estimation approaches are described in Sec.~\ref{chapter:mapping} and~\ref{sec:LVI-SLAM}. An experimental evaluation is given in Sec.~\ref{chapter:results}. 
\section{Related Work}
\label{chapter:related_work}
One of the first breakthroughs and still influential works in the field of LiDAR-based SLAM methods was LOAM~\cite{LOAM}. Its idea was the extraction of geometric features such as edges and planes, and matching them across frames. This basic concept has been widely adopted in many following works. While LOAM was also able to fuse IMU measurements in a loosely-coupled way, LIO-SAM~\cite{liosam} formulates a tightly-coupled pose graph optimisation fusing edge and plane error residuals with IMU error residuals. Also filter-based approaches, such as FAST-LIO~\cite{fastlio} or FAST-LIO2~\cite{fastlio2} have successfully demonstrated high-accuracy localisation fusing LiDAR and IMU. While FAST-LIO still uses also plane and edge features, FAST-LIO2 achieves significant speed-up by directly operating on the raw points.~\cite{balm} establishes sliding-window bundle adjustment (BA) for LiDAR scans.

Recently, LiDAR-Visual-Inertial SLAM systems have become increasingly popular. A large amount of these approaches makes use of LOAM's idea of geometric feature extraction. LVI-SAM~\cite{lvisam} combines two subsystems, a Visual-Inertial (VI) and a LiDAR-Inertial (LI), in a tightly-coupled way to complement each other in challenging scenarios. The LI system builds a factor graph based on IMU pre-integration errors and edge and plane residuals.
VILENS~\cite{Vilens} also builds an optimisation problem consisting of visual, inertial, leg odometry, and LiDAR-based line and plane residuals.   
Another line of research uses Multi-State Constraint Kalman Filter (MSCKF~\cite{msckf}) based approaches for tightly-coupled, filter-based fusion of visual, inertial and LiDAR measurements. Examples are~\cite{licfusion} and~\cite{licfusion2}. 

To minimise drift in long-term scenarios, the latest research commonly applies the concept of submapping. This originates from early SLAM research, such as the Atlas framework~\cite{atlas}.  
In this context, additional factors can be derived to align submaps and to eliminate drift. \cite{Vilens} uses local point-cloud submaps and adds ICP odometry measurements into the factor graph. Wildcat~\cite{Wildcat}, a sliding-window optimisation-based LiDAR-Inertial Odometry system, achieves peak state-of-the-art robustness and accuracy by building local surfel submaps and aligning the submaps in a pose graph optimisation. Most of the aforementioned work uses feature-based or surface-based 3D representations. While they might be sufficient for high-accuracy localisation and 3D reconstruction, they are not suitable for navigation due to their lack of representing observed free space.

Submapping on volumetric maps has been addressed in various works. \cite{ho2018virtual} builds occupancy submaps based on OctoMap~\cite{octomap} and aligns them by standard ICP registration. Voxgraph~\cite{voxgraph} uses a Visual-Inertial Odometry to provide poses for integration into TSDF maps. Upon completion, ESDF fields are generated and submaps are aligned in a back-end pose graph optimisation using correspondence-free error terms based on ESDF values. New submaps are created at a fixed frequency. Wang \textit{et al.}~\cite{oxfordLidarSE} instead use occupancy maps as their 3D representation. Using Supereight2~\cite{SE2} as adaptive-resolution mapping approach, new submaps are spawned based on the distance travelled. Submaps are re-arranged based on updates from the visual-inertial estimator. The follow-up work~\cite{oxfordSubmappingExtended} improved the submap creation by evaluating the point cloud overlaps of new scans and alignment of submaps is based on ICP. 
In this work, we will adopt the concept of submapping. In contrast to~\cite{voxgraph,oxfordLidarSE, oxfordSubmappingExtended}, the global alignment of submaps is not decoupled from the estimator but provides direct feedback. We formulate correspondence-free residuals as in~\cite{voxgraph} without the expensive need to extract ESDFs as we directly use the available occupancy information.

\section{Preliminaries}
\label{chapter:preliminaries}
\subsection{Problem Statement}
In this work, we aim to build an accurate and globally consistent volumetric representation of the environment around a mobile robot, which can be used in downstream tasks, most prominently navigation. 
We do so by taking a Visual-Inertial SLAM System, OKVIS2~\cite{OKVIS2}, as our starting point, as well as an adaptation of the adaptive-resolution occupancy mapping framework Supereight2~\cite{SE2}. As VI-SLAM tends to be locally consistent but suffers from the accumulation of larger drift over time, we tackle the problem of degrading map quality by applying submapping strategies. 
Furthermore, to increase the accuracy of the estimator while simultaneously ensuring global consistency of overlapping maps, we integrate LiDAR constraints based on frame-to-map as well as map-to-map alignment factors into the state estimator in a tightly-coupled way.
\subsection{Notation}
Throughout this work, the following notation will be used: coordinate frames are written as $\cframe{A}$ and a vector expressed in this reference frame will be denoted as $\pos{A}{}$. The rigid body transformation which transforms points from a reference frame $\cframe{B}$ to another reference frame $\cframe{A}$ is given by $\T{A}{B} \in SE(3)$ and can be decomposed into a rotation matrix $\mathbf{C}_{AB} \in SO(3)$ and the translational component $\pos{A}{B}$. We also denote the rotation $\mathbf{C}_{AB}$ with its unit quaternion form $\q{A}{B}$. 
The most important reference frames that will be used are: a fixed world reference frame $\cframe{W}$, the camera coordinate frames $\cframe{C_{i}}$ for $ i = 1\dots N$ cameras, the IMU sensor frame $\cframe{S}$, the LiDAR sensor frame $\cframe{L}$ and a map frame $\cframe{M}$.
Furthermore, $\crossmx{\cdot}$ represents the skew-symmetric matrix of a 3D vector.
\begin{figure}[t]
    \centering
    \hspace{-0.4cm}
    \includegraphics[width = 0.95\linewidth]{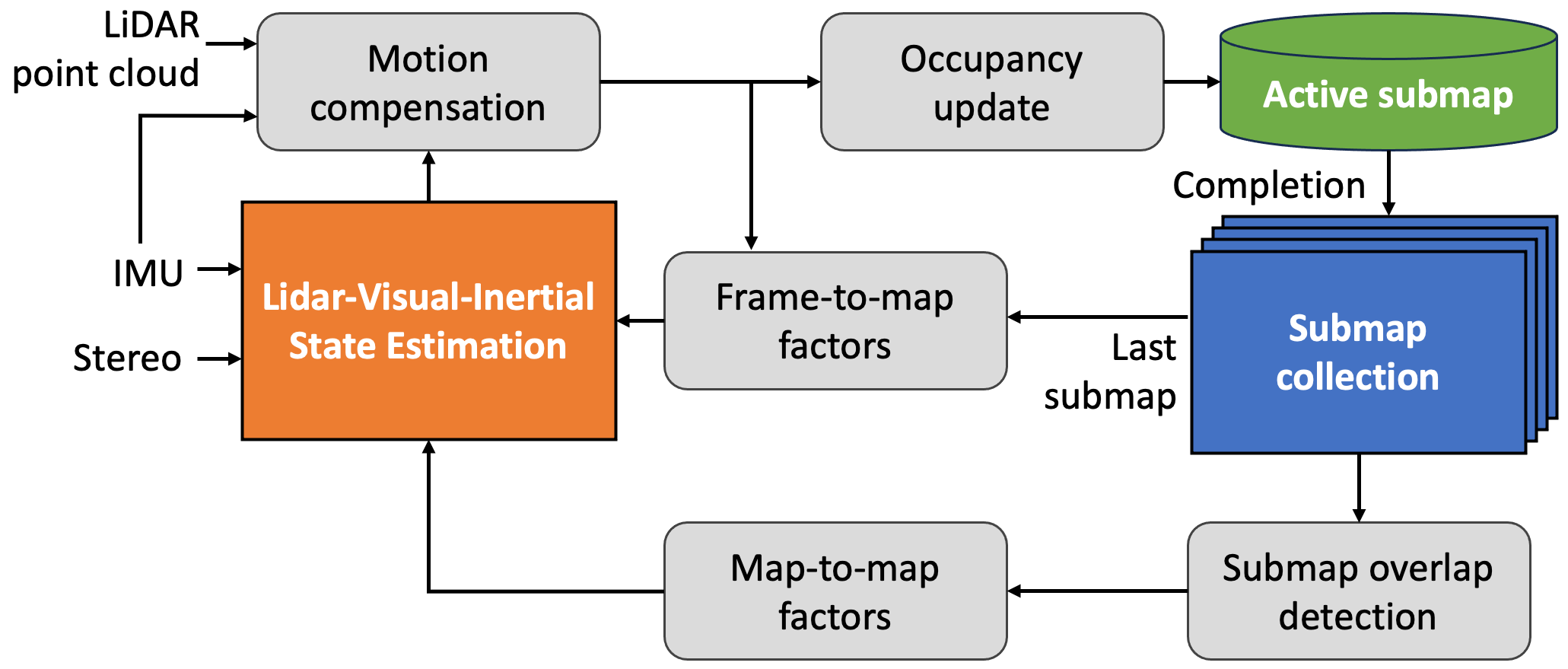}
    \caption{High-level structure of the proposed tightly-coupled LVI-SLAM and submapping framework. Motion compensation for LiDAR point clouds is performed to formulate live frame-to-map factors and to update the current active submap. Upon submap completion, the most overlapping previous submap is determined and map-to-map factors are fused into the optimisation problem.}
    \label{fig:system_overview}
    \vspace{-0.5cm}
\end{figure}
\section{System Overview}
\label{chapter:system_overview}
In this work, we build a fully tightly-coupled system combining volumetric mapping and LiDAR-Visual-Inertial (LVI) SLAM. A high-level overview is sketched in Fig.~\ref{fig:system_overview}. Inputs considered are IMU measurements, stereo camera frames and potentially unordered LiDAR point clouds. By the use of state estimates and IMU measurements integrated to a point's individual timestamp, motion compensation is performed to account for dynamic movement of the LiDAR sensor. After transforming the measurements to the respective map frame, the occupancy field of the current active submap is updated. Tight coupling between the submapping module and the LiDAR-Visual-Inertial estimator is achieved in two ways. Live frame-to-map factors are formulated for the state corresponding to the most recent stereo frame. Furthermore, upon completion of the active submap, the most overlapping submap is detected and map-to-map factors are added to the LVI state estimator. These factors as well as details regarding the submapping strategies will be given in the following sections. 
\begin{figure}[b]
    \centering
    \includegraphics[width=\linewidth]{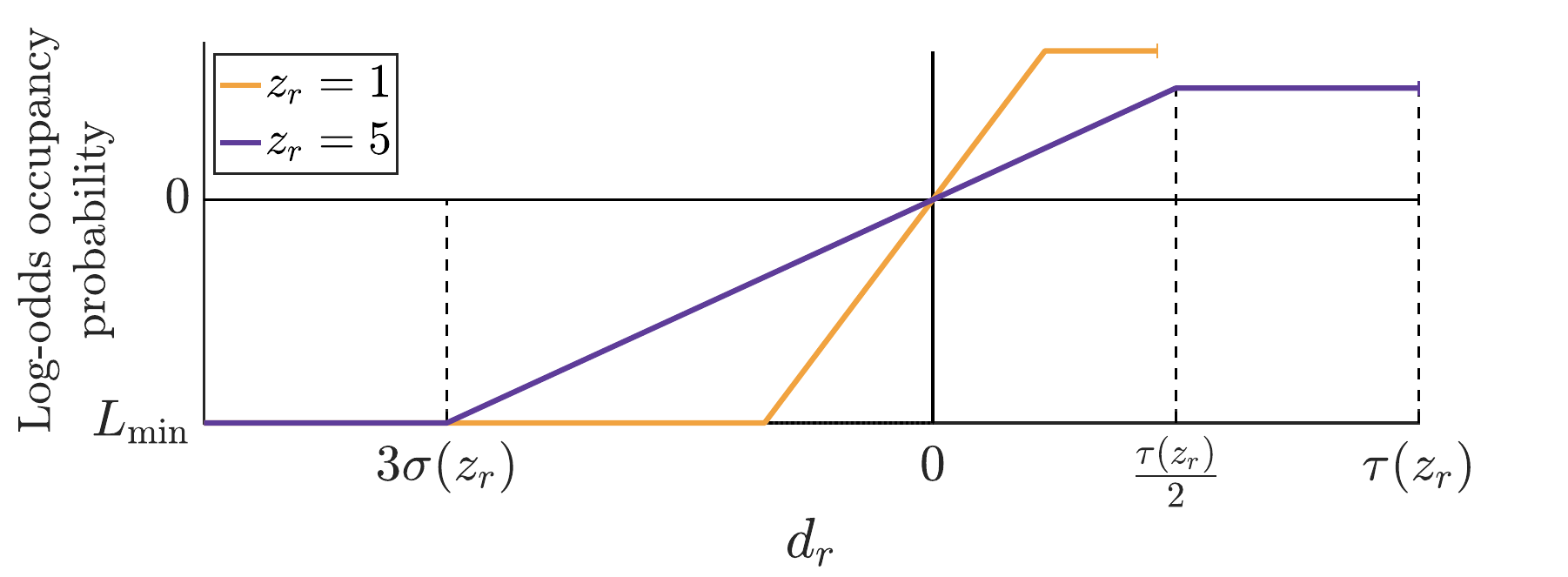}
    \caption{Inverse sensor model used in Supereight2~\cite{SE2} for two measurements (1m, 5m). The log-odds occupancy probability along  a ray measurement is expressed as a function of the difference $d_r$ between query points along the ray and the measured distance $z_r$. The occupancy values are clipped in front of the surface at a minimum $L_{\mathrm{min}}$ reached at $3 \sigma$ where $\sigma$ is a distance dependent uncertainty value. It grows linearly up to half the surface thickness $\tau(z_r)$. For more details, see~\cite{SE2}.
    \label{fig:inverse_sensor_model}}
    \vspace{-0.5cm}
\end{figure}
\section{Occupancy Mapping and Submapping}
\label{chapter:mapping}
As a volumetric occupancy mapping approach, we adopted the octree-based multi-resolution volumetric mapping framework Supereight2~\cite{SE2} specifically for the usage with different kinds of LiDAR sensors.
\subsection{Occupancy Mapping for dynamical LiDAR sensors}
In contrast to~\cite{oxfordLidarSE}, we do not use range image projections of LiDAR point clouds as input to our mapping system. Instead, to account for dynamically moving sensors, we added a ray-based integration interface for Supereight2 that fuses measurements on a per-ray-basis. We use fairly standard occupancy mapping in log-odds space with a simplified inverse sensor model that approximates the probabilistic nature of the sensor in terms of range or depth uncertainty and outlier probabilities. 
The log-odds occupancy $l^k \left( \mvec{M}{p} \right)$ of a 3D point $\mvec{M}{p}$ in a map frame $\cframe{M}$ at step $k$ is given by
\begin{equation}
    \label{eq:mapping_log_odd}
    l^k \left( \mvec{M}{p} \right) =
    \log \frac{P_{\mathrm{occ}} \left( \mvec{M}{p} | z_r \right)}{ 1 - P_{\mathrm{occ}} \left( \mvec{M}{p} | z_r \right)},
\end{equation}
where $z_r$ is the measured distance of a single ray and $l^k$ is following a piece-wise linear function along the ray as shown in Fig.~\ref{fig:inverse_sensor_model}.

Using clamped weights, we apply the same additive Bayesian updates as in~\cite{SE2}:
\begin{align}
    \label{eq:mapping_log_odd_update}
    \Bar{L}^{k} \left( \mvec{M}{p} \right) &= \frac{\Bar{L}^{k-1}\left( \mvec{M}{p} \right) w^{k-1} + l^k \left( \mvec{M}{p} \right)}{ w^{k-1} + 1}
    \nonumber \\ 
    w_k &= \min \left\{ w_{k-1} + 1, w_{\mathrm{max}}\right\}
\end{align}
The accumulated log-odds can still be preserved as
\begin{equation}
    \label{eq:log_odd_query}
    L^k \left( \mvec{M}{p} \right) = \Bar{L}^{k} \left( \mvec{M}{p} \right) w_k.
\end{equation}
To keep the octree representation consistent and the resolution as coarse as possible, similar up-propagation and tree pruning strategies as in~\cite{SE2} are applied.
\subsection{Submapping Strategy}
Similar to~\cite{oxfordSubmappingExtended}, our goal is to leverage the available sensor information as a decision criterion when to spawn a new submap: we evaluate per-frame the ratio between incoming measurements that correspond to already observed space and the total number of measurements. If this ratio falls below a certain threshold $\lambda_{\mathrm{overlap}}$, a new submap will be created with the next visual keyframe generated as in~\cite{OKVIS2}. Every submap is anchored in the IMU frame $\cframe{S}$ at the time step $k$ of the corresponding keyframe. All submaps are stored together with their respective keyframe poses $\T{W}{S^k}$.
\begin{figure*}[t]
    \centering
    \includegraphics[width=\linewidth]{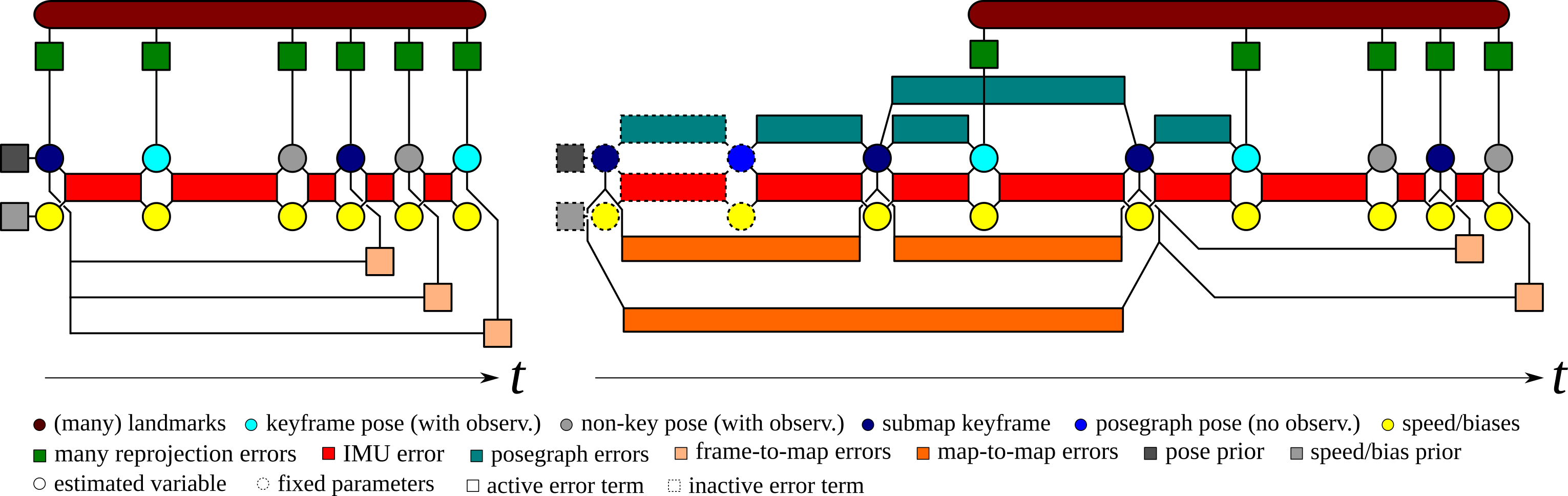}
    \caption{Optimisation Factor Graph of the LiDAR-Visual-Inertial Estimator. Left: The real-time estimator connects set of current keyframe states and non-keyframe states by IMU errors and visual reprojection errors. It also shows the active submap keyframe and the last completed submap keyframe. For every state in the optimisation window, we can formulate live LiDAR factors between every live frame and the last completed submap. Right: OKVIS2 connects keyframe poses through relative pose errors at a later stage. In addition to the live LiDAR factors, measurements between frames can be aggregated and map-to-map LiDAR factors can be added to the factor graph. Every map will be connected to the previous submap and optionally an older submap if the geometric overlap surpasses a threshold. }
    \label{fig:factor_graph}
    \vspace{-0.6cm}
\end{figure*}
\section{LiDAR-Visual-Inertial Estimator}
\label{sec:LVI-SLAM}
In the following, we present the extension of the OKVIS2~\cite{OKVIS2} factor graph. 
\subsection{Factor Graph and Optimisation Problem}
The factor graph representation used in this work is shown in Fig.~\ref{fig:factor_graph}. On top of visual, inertial and pose graph factors, that are already used in OKVIS2, two different types of LiDAR factors are added to the graph. For every state in the real-time optimisation window, LiDAR factors with respect to the last completed submap are added. Furthermore, map-to-map LiDAR factors are added to constrain submaps with respect to each other by aggregating measurements between submaps. 
The state vector that is estimated remains:
\begin{equation}
\label{eq:problemStatement-stateDef}
    \mathbf{x} = \left[ \pos{W}{S}^{T} , \q{W}{S}^{T}, \vel{W}[][]^{T}, \mathbf{b}_\mathrm{g}^{T} , \mathbf{b}_\mathrm{a}^{T} \right]^{T} ,
\end{equation}
where $\pos{W}{S}$ and $\q{W}{S}$ denote the position and orientation of the IMU sensor frame in the fixed world frame, and $\vel{W}[][]$ describes the velocity of the IMU with respect to the world frame. $\mathbf{b}_\mathrm{g}$ and $\mathbf{b}_\mathrm{a}$ stand for gyroscope and accelerometer biases, respectively.
\subsection{Error Residuals}
Here, for convenience, a summary of the original OKVIS2 factors is given before deriving the new LiDAR factors. 
\subsubsection{Visual Reprojection Errors}
OKVIS2~\cite{OKVIS2} uses 2D reprojection errors $ \mathbf{e}_{\mathrm{r}}^{i,j,k} $ of the $j$-th landmark in the frame of the $i$-th camera at a timestamp $k$ and its corresponding observation $\Tilde{\mathbf{z}}_{\mathrm{r}}^{i,j,k}$ in the image is given by
\begin{equation}
    \label{eq:residuals_reprojection}
    \mathbf{e}_{\mathrm{r}}^{i,j,k} 
    =
    \Tilde{\mathbf{z}}_{\mathrm{r}}^{i,j,k}  - \mathbf{h} \left( \mbfh{T}_{SC_i}^{-1} \T{S^k}{W} \leftidx{_W}{\mathbf{l}^{j}} \right) . 
\end{equation}
Hereby, $\mathbf{h} \left( \cdot \right)$ denotes the camera projection.
\subsubsection{IMU Errors}
For the formulation of the IMU residuals, OKVIS2 adopts the IMU preintegration approach in~\cite{forster2016manifold}. Between time steps $k$ and $n$, the IMU error is:
\begin{equation}
    \label{eq:residuals_imu}
    \mathbf{e}_{\mathrm{s}}^{k} 
    =
    \Tilde{\mathbf{x}}^{n} \left( \mathbf{x}^{k} , \Tilde{\mathbf{z}}_{\mathrm{s}}^{k,n}\right) \boxminus \mathbf{x}^{n} ,
\end{equation}
where $\Tilde{\mathbf{x}}^{n}$ is the predicted state at an arbitrary time $n$ as a function of the current state $ \mathbf{x}^{k} $ and IMU measurements $\Tilde{\mathbf{z}}_{\mathrm{s}}^{k,n}$. The $\boxminus$ performs regular subtraction except for the quaternion (see~\cite{OKVIS2}).
\subsubsection{Relative Pose Errors}
Additionally, relative pose errors $\mathbf{e}_{\mathrm{p}}^{r,c}$ between time steps $r$ and $c$ are given by 
\begin{equation}
    \label{eq:residuals_rel_pose}
    \mathbf{e}_{\mathrm{p}}^{r,c} 
    =
    \mathbf{e}_{\mathrm{p},0}^{r,c}
    +
    \begin{bmatrix}
      \pos{S^r}{S^c} - \postilde{S^r}{S^c} 
      \\
      \mathbf{q}_{S^r S^c} \boxminus \Tilde{\mathbf{q}}_{S^r S^c}
    \end{bmatrix}
    .
\end{equation}
with $\postilde{S^r}{S^c}$ and $\Tilde{\mathbf{q}}_{S^r S^c}$ being nominal relative position and orientations expressed in the IMU frame $\cframe{S}$. We refer the reader to~\cite{OKVIS2} for a detailed derivation of the constant $\mathbf{e}_{\mathrm{p},0}^{r,c}$ as well as error Jacobians and error weights $ 
\mathbf{W}_\mathrm{p}^{r,c} $.
\subsubsection{Submap-based LiDAR Errors}
As stated in the previous section, there are two types of LiDAR residuals.
In both cases, the error residuals are formulated based on two states. On the one hand, for live residuals, these two states are given by the current frame and the keyframe associated to the last completed submap. On the other hand, for map-to-map residuals, both states will be submap anchor frames.
Regarding the formulation of the error residuals, however, they do not differ.

Given a completed submap 
anchored at keyframe pose $\T{W}{S^a}$ and a point cloud $\mathcal{P}_{b}$ associated to another state $\T{W}{S^b}$, we formulate the residual for every $\mvec{S^b}{p} \in \mathcal{P}_{b}$ as:
\begin{equation}
    \label{eq:method_lidar_residual}
    e^{a,b}_{\mathrm{l}} \left( \mvec{S^a}{p} \right) =
    \frac{d}{\sigma} 
    = 
    \frac{L\left( \mvec{S^a}{p} \right)}
    {\sqrt{\frac{L_{\mathrm{min}}^{2}}{9} + \sigma_{z}^{2} \left| \nabla L \left( \mvec{S^a}{p} \right)\right|^{2}}} ,
\end{equation}
where $\mvec{S^a}{p} = \T{S^a}{S^b} \mvec{S^b}{p}$ with $\T{S^a}{S^b} = \T{W}{S^a}^{-1} \T{W}{S^b}$. The idea here is that every measured point should be on a surface in the 3D map; and the distance $d$ of the point from the nearest surface can be extrapolated from the occupancy value $L\left( \cdot \right)$ and the occupancy gradient $\nabla L \left( \cdot \right) $ assuming a linear behavior as in the sensor model for mapping. From the model (Fig.~\ref{fig:inverse_sensor_model}), we can derive the distance $d$ and the map uncertainty as:
\begin{equation}
    \label{eq:method_map_distance_uncertainty}
    d = \frac{L}{|\nabla L|}, \qquad
    \sigma_{\mathrm{map}} = \frac{L_{\mathrm{min}}}{3 | \nabla L |} .
\end{equation}
$L_{\mathrm{min}}$ is a configuration parameter and denotes the saturation minimum log-odds occupancy value.
With the sensor-specific measurement uncertainty $\sigma_{z}$, we can formulate the weighted residual in Eqn.~\eqref{eq:method_lidar_residual} using the total uncertainty
\begin{equation}
    \label{eq:method_total_uncertainty}
    \sigma = \sqrt{\sigma_{\mathrm{map}}^{2} + \sigma_{z}^{2}}.
\end{equation}
Furthermore, we want to derive analytic Jacobians, whereby using a perturbation $\delta \mbf{\chi}_T = [\delta \mbf{r}, \delta \mbf{\alpha}]$ for poses around linearisation points $\mbfbar{r}$ and $\mbfbar{C}$:
\begin{equation}
\label{eq:error-state}
    \begin{aligned}
    \mbf{r} &= \bar{\mbf{r}} + \delta \mbf{r}, \\
    \mbf{C} &= \Exp {\delta \boldsymbol{\alpha}} \mbfbar{C}.
    \end{aligned}
\end{equation} 
We further define the error state as
$\delta \boldsymbol{\chi} = [\delta \mbf{\chi}_{T_{WS}}, \delta\mbf{\chi}_\mathrm{sb}]$, with $\delta\mbf{\chi}_\mathrm{sb}$ denoting an additive perturbation of the speed and biases. With the LiDAR residuals only depending on pose states, but not on speed and biases, we can compute Jacobians with respect to the poses of frames $\cframe{S^{a}}$ and $\cframe{S^{b}}$ leveraging the chain rule:
\begin{equation}
    \label{eq:jacobian_chain_rule}
    \frac{\partial e^{a,b}_{\mathrm{l}}}{\delta \mbf{\chi}_{T_{WS^m}}} 
    = 
    \frac{\partial e^{a,b}_{\mathrm{l}}}{\partial \mvec{S^a}{p}}
    \frac{\partial \mvec{S^a}{p}}{\delta \mbf{\chi}_{T_{WS^m}}},
\end{equation}
with $m \in \{ a, b \}$. The individual Jacobians in Eq.~\eqref{eq:jacobian_chain_rule} are:
\begin{align}
    \label{eq:jacobians}
    \frac{\partial e^{a,b}_{\mathrm{l}}}{\partial \mvec{S^a}{p}} &= 
    \frac{\nabla L}{\sqrt{\frac{L_{\mathrm{min}}^{2}}{9} + \left| \nabla L \right|^{2}\sigma_{z}^{2}}}
    \nonumber \\
    \frac{\partial \mvec{S^a}{p}}{\delta \mbf{\chi}_{T_{WS^a}}} &= \C{S^a}{W}
    \begin{bmatrix}
      -\mbf{I}_3 & \crossmx{{\C{W}{S^b}}^\mvec{S^b}{p} + \pos{W}{S^b} - \pos{W}{S^a}}
    \end{bmatrix}
    \nonumber \\
    \frac{\partial \mvec{S^a}{p}}{\delta \mbf{\chi}_{T_{WS^b}}} &= \C{S^a}{W}
    \begin{bmatrix}
      \mbf{I}_3 & - \crossmx{{\C{W}{S^b}}\mvec{S^b}{p}}
    \end{bmatrix}
    .
\end{align}

Note that the overall idea of this optimisation resembles a point-to-plane ICP. In contrast to the usual point-to-plane ICP, the expensive step of data association and normal computation can be omitted as occupancy values and gradients can be directly retrieved from the occupancy field. Furthermore, outliers are implicitly covered as they will most likely lie in either free or unobserved space with zero or invalid gradients.
\subsubsection{Optimisation Problem}
All of the aforementioned factors are combined in the overall minimisation objective:
\begin{align}
    \label{eq:optimization_objective}
    c\left( \mathbf{x} \right) &= 
    \frac{1}{2} \sum_{i} \sum_{k \in \mathcal{K}} \sum_{j \in \mathcal{J} \left(i,k \right)} \rho \left( {\mathbf{e}_{\mathrm{r}}^{i,j,k}}^{T} \mathbf{W}_{\mathrm{r}} \mathbf{e}_{\mathrm{r}}^{i,j,k}   \right)
    \nonumber \\
    &+ \frac{1}{2} \sum_{k \in \mathcal{P} \cup \mathcal{K} \setminus f} 
    {\mathbf{e}_{\mathrm{s}}^{k}}^{T} \mathbf{W}_\mathrm{s}^{k} \mathbf{e}_{\mathrm{s}}^{k} 
    + \frac{1}{2} \sum_{r\in \mathcal{P}} \sum_{c\in \mathcal{C}\left(r\right)} {\mathbf{e}_{\mathrm{p}}^{r,c}}^{T} \mathbf{W}_{\mathrm{r}}^{r,c} \mathbf{e}_{\mathrm{p}}^{r,c}   
    \nonumber \\
    &+ \frac{1}{2} \sum_{k \in \mathcal{K}}\sum_{\mathbf{p} \in \mathcal{L}_k}{{e_{\mathrm{l}}^{C,k}}^2}
    + \frac{1}{2} \sum_{b \in \mathcal{M}}\sum_{a \in \mathcal{A}_b}\sum_{\mathbf{p} \in \mathcal{L}_b}{{e_{\mathrm{l}}^{a,b}}^2}.
\end{align}
Here, the set $\mathcal{K}$ contains the most recent frames as well as keyframes with observations of visible landmarks in $\mathcal{J} \left(i,k \right)$. $\mathcal{P}$ contains all pose graph frames and $f$ denotes the most current frame. $\mathcal{C}\left(r\right) \subset \mathcal{P}$ is the set of all pose graph frames connected to a frame $r$. 
Furthermore, the set $\mathcal{L}_k$ denotes the set of all LiDAR measurements associated to a frame $k$. $\mathcal{M}$ is the set of all past submaps, and $\mathcal{A}_b$ the set of all submap frames connected to a submap frame $\T{W}{S^b}$ via map-to-map residuals. $C$ denotes the last completed submap frame.

\setlength{\tabcolsep}{2.5pt}
\begin{table*}[tb]
\begin{center}
\begin{tabular}{l |c| c c c | c  c  c  c  c  c  c  c  c  c  c  | c }
\Xhline{3\arrayrulewidth}
\textbf{Approach} & & \multicolumn{3}{c|}{\textbf{Sensors}} & \multicolumn{11}{c|}{\textbf{Sequence}} & \textbf{Score} \\
 & C & L & V & I & exp01(e) & exp02(m) & exp03(h) & exp04(e) & exp05(e) & exp06(m) & exp07(m) & exp09(h) & exp11(m) & exp15(h) & exp21(e) \\
\Xhline{3\arrayrulewidth}
\multirow{1}{*}{OKVIS2~\cite{OKVIS2}} & \cmark &  & \cmark & \cmark & \multirow{1}{*}{8.46} & \multirow{1}{*}{8.18} & \multirow{1}{*}{0.00} & \multirow{1}{*}{\textcolor{gray}{17.14}} & \multirow{1}{*}{\textcolor{gray}{23.33}} & \multirow{1}{*}{\textcolor{gray}{14.29}} & \multirow{1}{*}{0.00} & \multirow{1}{*}{0.00} & \multirow{1}{*}{0.00} & \multirow{1}{*}{0.00} & \multirow{1}{*}{0.00} & \multirow{1}{*}{16.64} 
\\
\multirow{1}{*}{OKVIS2~\cite{OKVIS2}} & &  & \cmark & \cmark & \multirow{1}{*}{30.77} & \multirow{1}{*}{20.00} & \multirow{1}{*}{0.00} & \multirow{1}{*}{\textcolor{gray}{47.14}} & \multirow{1}{*}{\textcolor{gray}{56.67}} & \multirow{1}{*}{\textcolor{gray}{\underline{34.28}}} & \multirow{1}{*}{\underline{31.67}} & \multirow{1}{*}{13.75} & \multirow{1}{*}{28.00} & \multirow{1}{*}{12.22} & \multirow{1}{*}{0.00} & \multirow{1}{*}{136.41} 
\\
\hline
\multirow{1}{*}{VILENS~\cite{Vilens}} & & \cmark & \cmark & \cmark & \multirow{1}{*}{69.23} & \multirow{1}{*}{49.09} & \multirow{1}{*}{40.00} & \multirow{1}{*}{-} & \multirow{1}{*}{-} & \multirow{1}{*}{-} & \multirow{1}{*}{16.67} & \multirow{1}{*}{5.00} & \multirow{1}{*}{68.00} & \multirow{1}{*}{17.78} & \multirow{1}{*}{60.00} & \multirow{1}{*}{325.77} 
\\
\multirow{1}{*}{HKU$^*$} & \cmark & \cmark & \cmark & \cmark & \multirow{1}{*}{\textbf{84.62}} & \multirow{1}{*}{\underline{70.00}} & \multirow{1}{*}{\underline{44.71}} & \multirow{1}{*}{-} & \multirow{1}{*}{-} & \multirow{1}{*}{-} & \multirow{1}{*}{\underline{31.67}} & \multirow{1}{*}{7.50} & \multirow{1}{*}{\textbf{92.00}} & \multirow{1}{*}{24.44} & \multirow{1}{*}{48.00} & \multirow{1}{*}{402.93}
\\
\multirow{1}{*}{Wildcat~\cite{Wildcat}} & & \cmark &  & \cmark & \multirow{1}{*}{\textbf{84.62}} & \multirow{1}{*}{\textbf{84.55}} & \multirow{1}{*}{\textbf{90.59}} & \multirow{1}{*}{-} & \multirow{1}{*}{-} & \multirow{1}{*}{-} & \multirow{1}{*}{\textbf{45.00}} & \multirow{1}{*}{\textbf{44.38}} & \multirow{1}{*}{84.00} & \multirow{1}{*}{\underline{46.67}} & \multirow{1}{*}{\textbf{84.00}} & \multirow{1}{*}{\textbf{563.79}} 
\\
\multirow{1}{*}{\textbf{Ours}} & \cmark & \cmark & \cmark & \cmark & \multirow{1}{*}{43.08} & \multirow{1}{*}{17.27} & \multirow{1}{*}{32.35} & \multirow{1}{*}{\textcolor{gray}{\underline{57.14}}} & \multirow{1}{*}{\textcolor{gray}{\underline{61.67}}} & \multirow{1}{*}{\textcolor{gray}{17.14}} & \multirow{1}{*}{0.00} & \multirow{1}{*}{16.25} & \multirow{1}{*}{32.00} & \multirow{1}{*}{51.11} & \multirow{1}{*}{62.00} & \multirow{1}{*}{254.06}
\\
\multirow{1}{*}{\textbf{Ours}} & & \cmark & \cmark & \cmark & \multirow{1}{*}{62.31} & \multirow{1}{*}{39.09} & \multirow{1}{*}{42.35} & \multirow{1}{*}{\textcolor{gray}{\textbf{68.57}}} & \multirow{1}{*}{\textcolor{gray}{\textbf{86.67}}} & \multirow{1}{*}{\textcolor{gray}{\textbf{47.14}}} & \multirow{1}{*}{13.33} & \multirow{1}{*}{\underline{38.12}} & \multirow{1}{*}{\textbf{92.00}} & \multirow{1}{*}{\textbf{64.44}} & \multirow{1}{*}{\underline{76.00}} & \multirow{1}{*}{\underline{427.65}}  \\
\hline
\end{tabular}
\end{center}
\vspace{-0.25cm}
\caption{Evaluation scores on HILTI22 SLAM Challenge. Bold: best score, underlined: second best; C: causal evaluation (definition in~\cite{OKVIS2}); L, V, I: LiDAR, Visual and Inertial measurements are considered; (e/m/h) classifies the difficulty level of the sequence: easy, medium or hard; $^{*}$based on~\cite{fastlivo}. Sequences exp04, exp05 and exp06 greyed out as they are not included in the final score.}
\label{tab:hilti22}
\vspace{-0.5cm}
\end{table*}
\section{Experimental Results}
We will quantitatively evaluate the accuracy of the proposed tightly-coupled LiDAR-Visual-Inertial SLAM system in Section~\ref{sec:Results_HILTI}. In section~\ref{sec:Results_Mapping}, we will qualitatively demonstrate that the proposed system yields globally consistent submaps that can be used in robotic applications. Finally, in Section~\ref{sec:Results_Timing}, we will briefly analyse the computational performance of the proposed system.
\label{chapter:results}
\subsection{Implementation Details}
All experiments were performed on an Intel i7-11700K CPU with 3.6 GHz and 64 GB RAM.
Google's non-linear least squares optimisation framework Ceres~\cite{Ceres} has been used with an implementation of analytical Jacobians. 
As a threshold $\lambda_{\mathrm{overlap}}$ for new submap creation, a value of $0.4$ was chosen. For every live state, we added $100$ frame-to-map residuals. Upon completion of submaps, $1000$ map-to-map residuals were added. For that, points were randomly sampled from the aggregated point clouds between the two corresponding frames. Submaps have a maximum dimension of $15.36$ m and a finest resolution of $3$ cm.
\subsection{HILTI Slam Challenge}
\label{sec:Results_HILTI}
We evaluated our approach on the HILTI 2022 SLAM Challenge benchmark~\cite{Hilti22}. The Hilti-Oxford dataset was collected using a handheld sensor device consisting of an Alphasense five-camera module, an IMU and a 32 beam Hesai PandarXT-32 LiDAR sensor. With the available calibration datasets, we ran our own sensor calibration of the multi-camera system and IMU-camera extrinsics using Kalibr~\cite{Kalibr}. As we experienced that the performance on the different sequences dependent significantly on the calibration parameters that were used, we also performed online calibration of the IMU-camera extrinsics. The challenge includes 8 different sequences with varying levels of difficulty. Difficulty levels are based on challenging environments (e.g.\ bad lighting) or aggressive motions of the sensor suite. Out of the 8 sequences, 3 were taken on a construction site (\textit{Exp01}, \textit{Exp02} and \textit{Exp03}), 1 was taken in a long corridor of an office building (\textit{Exp07}) and the remaining 4 sequences were recorded in and around the Oxford Sheldonian Theatre (\textit{Exp09}, \textit{Exp11}, \textit{Exp15} and \textit{Exp21}). After the challenge, 3 additional sequences from the construction site (\textit{Exp04}, \textit{Exp05} and \textit{Exp06}) were added to the automatic evaluation system.
For evaluation, sparse ground-truth with millimeter accuracy is provided. An automatic evaluation system is available that uses a score based on the Absolute Trajectory Error (ATE) as an evaluation metric. First, the estimated trajectories are aligned with the sparse ground-truth using $SE(3)$ Umeyama alignment. For every ground-truth control point, the corresponding estimate is retrieved and awarded a score ranging from $0$ to $10$ ($10$ for errors below $1$ cm, $0$ for errors above $10$ cm, for more details see~\cite{Hilti22}). The total score $S_j$ for a dataset $j$ with $N$ control points is computed as
\begin{equation}
    \label{eq:hilti_score}
    S_j = \left( \frac{1}{10N} \sum_{i=0}^{N} s_i \right) \times 100. 
\end{equation}
A full score of 100 would correspond to a localisation error below 1 cm across the whole trajectory. 

Table~\ref{tab:hilti22} shows the final scores of our approach on all challenge sequences with identical parameters. We report the causal as well as non-causal estimates including a final Bundle Adjustment. We compare it to the Visual-Inertial SLAM (including loop closures) performance of OKVIS2~\cite{OKVIS2} as a baseline. The VI baseline used all 5 cameras and the same visual frontend parameters as in the LVI case. Furthermore, we compare it to other state-of-the-art methods that have been published an evaluated on the benchmark including the -- at the time of the challenge -- winning algorithm Wildcat~\cite{Wildcat}.

It can be seen that adding LiDAR factors increases the accuracy of OKVIS2 significantly, in the causal as well as the final optimised evaluation. Furthermore, the proposed approach achieves state-of-the-art localisation accuracy. At the moment, a final score of 427.65 achieves rank 7 out of 53 submissions on the challenge leaderboard. Note that also VILENS and Wildcat performed some form of offline batch or posegraph optimisation for the reported scores. For a better understanding, it can be stated that the reported scores here denote a mean ATE of 1 cm (or even below) to 3 cm in the final estimates. Only for \textit{exp07} where a long corridor with a lack of characteristic visual features or geometric features does not achieve the same performance and results in a mean ATE of approximately 7 cm. 
\begin{figure}[b]
    \centering
    \includegraphics[width = 1\linewidth, trim={0.8cm, 0.4cm, 0.8cm, 0.4cm}, clip]{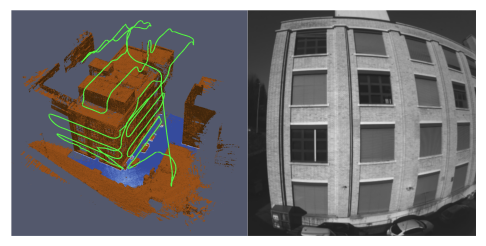}
    \caption{Outdoor Example: Reconstruction (brown), horizontal slice of the free space (blue) and estimated trajectory (green).}
    \label{fig:leica_example}
\end{figure}
\subsection{Mapping}
\label{sec:Results_Mapping}
\subsubsection{HILTI Dataset}
Fig.~\ref{fig:reconstruction} shows an overlay of all submaps created on \textit{Exp21}. Meshes of the surfaces and a slice through the occupancy fields at a height of $0.2$ m are shown. It qualitatively demonstrates that we are able to provide accurate 3D reconstructions and sensible occupancy fields immediately usable for robot navigation without discernible inconsistencies between submaps.
\subsubsection{Real-World Example}
To qualitatively demonstrate the approach's versatility to a variety of different LiDAR sensor types, we additionally processed a dataset recorded on a Leica BLK2Fly drone. This dataset has been recorded in the outdoor area around an office building. The drone is equipped with five cameras and also five IMUs, out of which we only use one for state estimation. The LiDAR sensor is a single-beam dual-axis spinning LiDAR sensor. Our mapping approach avoids projection into very sparse range images. Fig.~\ref{fig:leica_example} shows a reconstruction as well as the trajectory and a slice through the occupancy field.
\subsection{Timings}
\label{sec:Results_Timing} 
We further show timings for map integration and graph optimisation of \textit{Exp15} in Table~\ref{tab:timings}.
LVI Setup 1 here denotes the original setup as it was used for the evaluation of all sequences in~\ref{tab:hilti22}. In Setup 2, we reduced the number of LiDAR residuals to $50$ for live frame-to-map and $500$ for map-to-map factors. Furthermore, the input point clouds were downsampled with a factor of $3$. The results show that the rate of the optimisation can be increased from approximately $10 $ Hz to almost $15$ Hz. Also the map integration can be sped up significantly by downsampling input point clouds for mapping. Instead of running at $10$ Hz, we can run integration at $25$ Hz. Note, that this speed-up only leads to a very minor degradation in terms of localisation accuracy.
\begin{table}[h]
    \centering
    \begin{tabular}{c c c c}
    \textbf{Function}  & \textbf{VI only}& \textbf{LVI Setup 1} & \textbf{LVI Setup 2} \\
    \hline
    Optimisation & $ 49.0 \pm 16.6$ & $ 98.9 \pm 30.9 $ & $79.3 \pm 25.0$ \\
    Batch Integration & - & $ 98.4 \pm 83.0$ & $38.2 \pm 34.5$ \\
    \textbf{ATE (Causal) [cm]} & $23.5$ & $2.5$ & $3.8$ \\
    \hline
    \end{tabular}
    \caption{Per-frame average timings in ms and standard deviations for optimisation and map integration on \textit{Exp15}.}
    \label{tab:timings}
\end{table}
\section{Conclusion}
\label{chapter:conclusion}
Autonomous navigation in the real-world requires high-accuracy localisation, but also accurate and more importantly a globally consistent 3D representation of the environment. In this paper, we propose a fully tightly-coupled LiDAR-Visual-Inertial system that addresses both of these tasks simultaneously. Results from the state estimator are used to integrate incoming LiDAR measurements into local submaps. A novel, correspondence-free residual formulation has been introduced that only uses occupancy field values and gradients which can be efficiently queried. In a tightly-coupled approach, these LiDAR factors can be added to the optimisation problem either as live frame-to-map constraints or as map-to-map constraints across larger time spans. The proposed system has proven to achieve state-of-the-art performance in localisation while yielding consistent occupancy submaps. In future work, we are planning to improve on several aspects. These include amongst others a more realistic uncertainty model for the LiDAR sensor or an informed way of downsampling incoming LiDAR scans leveraging e.g. geometric characteristics. Another challenge, that we will address in the future is the robustness to challenging scenarios in which the visual frontend is prone to tracking failures. 
Finally, we are also planning to extend the current work to a full exploration framework navigating through submaps in the real-world.


{
\bibliographystyle{IEEEtran}
\bibliography{root}
}

\end{document}